\documentclass{article}

\usepackage{PRIMEarxiv}
\usepackage[utf8]{inputenc}
\usepackage[T1]{fontenc}
\usepackage{graphicx}
\usepackage{amsmath}
\usepackage{amssymb}
\usepackage{amsfonts}
\usepackage{booktabs}
\usepackage[table]{xcolor}
\usepackage{multirow}
\usepackage{microtype}
\usepackage{xspace}
\usepackage[breaklinks,colorlinks,citecolor=blue,linkcolor=blue,urlcolor=blue]{hyperref}
\usepackage[nameinlink,capitalise]{cleveref}
\makeatletter
\let\oldthebibliography\thebibliography
\renewcommand{\thebibliography}[1]{%
  \oldthebibliography{#1}%
  \footnotesize
  \setlength{\itemsep}{0pt}%
  \setlength{\parskip}{0pt}%
  \setlength{\parsep}{0pt}%
  \sloppy
}
\makeatother

\graphicspath{{./}}

\begin{document}

\title{Personalizing LLM Agent Memory Using Biometrics}
\author{
Yanhong Qian, Qingguo Meng, Shihao Ding, Xingbo Dong, Zhe Jin,\\
Hanrui Wang, and Isao Echizen (Senior Member, IEEE)\\
Anhui Provincial Key Laboratory of Secure Artificial Intelligence,\\
School of Artificial Intelligence, Anhui University, Hefei 230093, China\\
\texttt{yanhongqian@stu.ahu.edu.cn; mqg1024@163.com;}\\
\texttt{shihaoding@stu.ahu.edu.cn; xingbo.dong@ahu.edu.cn;}\\
\texttt{jinzhe@ahu.edu.cn}\\[3pt]
National Institute of Informatics, Tokyo 101-8430, Japan\\
\texttt{hanrui\_wang@nii.ac.jp; iechizen@nii.ac.jp}
}
\maketitle

\begin{abstract}
Personalized memory helps LLM agents deliver stable, tailored assistance by storing and reusing user-specific data across interactions. In multi-user scenarios, however, retrieval must consider not only semantic similarity but also whether the current requester matches the identity associated with the stored memory. We propose Bio-Memory, a biometric-aware memory architecture that conditions memory retrieval on both semantic similarity and biometric matching. Built on top of A-Mem, Bio-Memory augments each atomic memory note with a biometric embedding and uses biometric matching to form the retrieval candidate pool before semantic ranking. We evaluate Bio-Memory on LoCoMo in a 10-user shared-agent setting over 7 face benchmarks and 10 palmprint protocols. Across datasets, Bio-Memory consistently separates owner and non-owner queries. Under face-based personalization, the largest average gap reaches 27.29\% / 21.15\% in F1 / BLEU-1 on CALFW; under palmprint-based personalization, the corresponding gap is 25.75\% / 19.22\% on MS\_Blue. These results support biometrics as a practical control signal for personalized memory retrieval in shared environments.
\keywords{face recognition \and palmprint recognition \and identity-aware memory \and multi-user LLM agents}
\end{abstract}

\section{Introduction}
%先说目前的agent需要存储个人的信息
Large language model (LLM) agents are increasingly expected to function as persistent personalized assistants rather than one-shot question answering systems \cite{zhang2026generalizability}. To support long-term interaction, recent memory architectures allow agents to store and reuse user-related information such as preferences, schedules, historical interactions, and task-specific facts across sessions \cite{du2025rethinking}. A representative example is A-Mem \cite{xu2025mem}, which organizes memories as structured notes containing fields such as timestamp, content, keywords, context, tags, and semantic embeddings. More broadly, systems such as MemoryBank \cite{zhong2024memorybank}, Mem0 \cite{singh2025mem0} and PersonaMem-v2 \cite{jiang2025personamem} show that memory can substantially improve interaction continuity and user adaptation \cite{du2025rethinking}.

%但是目前agent开始偏向多用户同时用一个agent了，那就会带来新的问题，之前的账号的形式是没有办法确认使用该对话的是同一个人呀
However, as memory-augmented LLM agents move from single-user devices to shared environments, personalization brings a new challenge: the system must not only retrieve relevant memories, but also retrieve the \emph{right user's} memories. Existing personalized memory systems are generally built around an externally supplied identity, such as an account, user ID, or session boundary \cite{du2025rethinking}. This assumption is fragile in shared settings, where a smart home assistant, public terminal, or enterprise assistant may serve multiple people via the same device or active session. Accounts, passwords, and session states can be shared, reused, or left active, so identifying the active profile differs from identifying the current human requester. For personalized memory retrieval, this distinction is critical: the same query, such as ``What is my next appointment?'', should retrieve different memories depending on who is physically making the request.

\begin{figure}[t]
\begin{center}
   \includegraphics[width=\columnwidth]{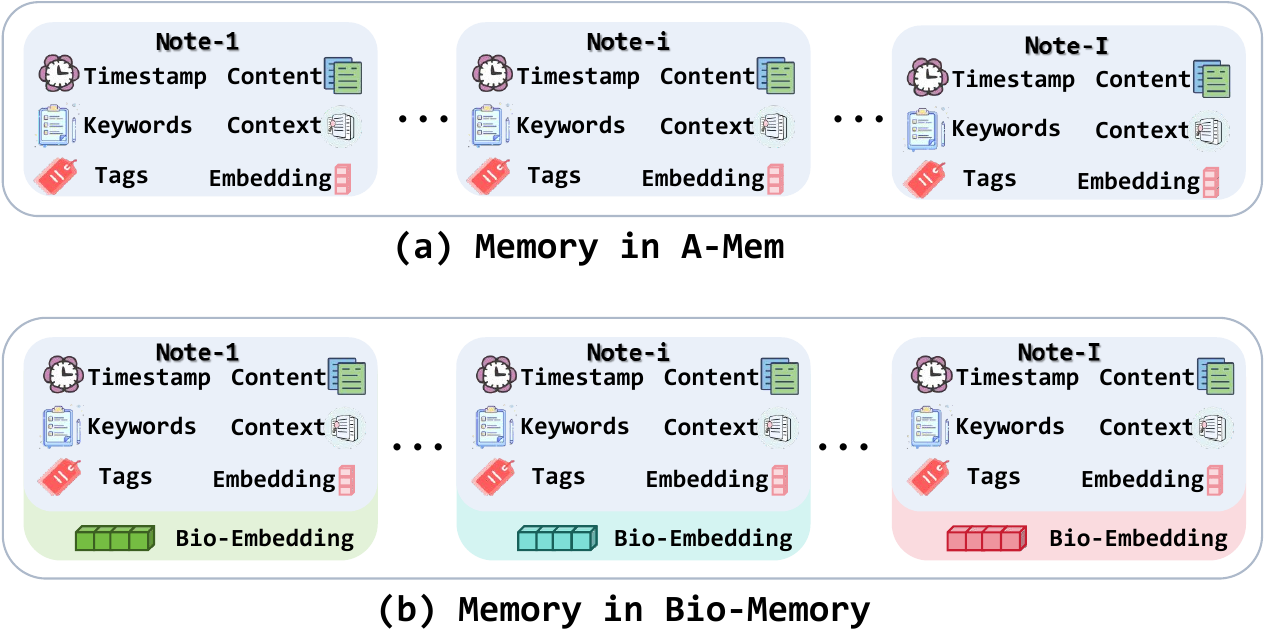}
\end{center}
\caption{Comparison of memory note structures in (a) A-Mem and (b) the proposed Bio-Memory. The atomic memory note of A-Mem consists of timestamps, content, keywords, context, tags, and semantic embeddings. Bio-Memory inherits this structure and additionally integrates biometric embeddings. Different colors in (b) denote memories from distinct users, supporting joint semantic and biometric matching for accurate memory retrieval in shared scenarios.}
\label{fig:intro}
\end{figure}

%讲我们是怎么做的
Figure \ref{fig:intro} shows the structural difference between A-Mem and Bio-Memory. Bio-Memory preserves the semantic fields of atomic memory notes while adding a biometric embedding that ties each note to a physical user. This changes personalization from purely semantic retrieval to retrieval conditioned on both semantic relevance and the current user's biometric identity. Rather than using biometrics only for session authentication, we use them to determine which memories may enter the candidate pool. This design is particularly useful in shared environments, where the session state may remain unchanged even when the active requester changes.

%讲我们的方法
Based on this idea, we propose \textbf{Bio-Memory}, a biometric-aware memory architecture built on top of A-Mem. Bio-Memory preserves the atomic note organization of A-Mem while augmenting each note with an additional biometric embedding. At inference time, the current user's probe biometric embedding is matched against the stored biometric embedding attached to each memory note, and semantic retrieval is then performed only over the resulting biometric-matched candidate memories. We evaluate Bio-Memory in a 10-user shared-agent setting using LoCoMo together with public face and palmprint benchmarks.

Our contributions are threefold: i) we augment atomic memory notes with biometric identity evidence, grounding agent memories in both semantic content and the physical identity of the requester; ii) we propose Bio-Memory, a biometric-aware extension of A-Mem that constructs a biometric-matched candidate pool before semantic retrieval and response generation; iii) we present an extensive 10-user shared-agent evaluation on LoCoMo, covering seven face benchmarks and ten palmprint protocols.

\section{Related Work}
\label{sec:related-work}
\subsection{Memory-Augmented LLM Agents}
%先讲长期记忆
Long-interaction LLM agents rely on persistent external memory to retain user state beyond the limited context window of a single prompt \cite{yehudai2026survey,yao2025collaborative}. Early memory-augmented agent frameworks primarily target long-term dialogue continuity. Generative Agents records all user observations into a textual memory stream, ranks stored records via recency, relevance and importance, and periodically generates high-level reflective summaries to support downstream planning and decision-making \cite{park2023generative}. To resolve identical context-length bottlenecks, MemGPT adopts an OS-inspired memory architecture. It regards the prompt as finite working memory and implements explicit memory paging to swap content between in-context windows and external storage \cite{packer2024memgpt}. Collectively, these pioneer works establish two foundational design principles for agent memory. Personal history must be decoupled from the prompt context, and retrieval rules dominate which historical information enters model reasoning.

Subsequent research evolves generic long-term memory toward user-personalized storage. MemoryBank equips dialogue agents with dedicated long-term memory, event summaries and static user profiles, and introduces an Ebbinghaus-inspired forgetting mechanism to dynamically adjust memory activation levels instead of treating all records as equally persistent \cite{zhong2024memorybank}. Treating memory management as a data maintenance pipeline, Mem0 automatically extracts core factual knowledge from conversations, consolidates sparse records into compact storage, and supports standard CRUD operations to maintain memory consistency. Its graph-enhanced extension further models relational dependencies between stored facts \cite{singh2025mem0}. PersonaMem-v2 explores implicit user personalization across disjoint multi-turn sessions, demonstrating that compactly compressed user memory drastically cuts token overhead compared to feeding full dialogue history, while preserving strong personalized reasoning performance \cite{jiang2025personamem}. Personalized dialogue agents push this direction further. LD-Agent combines long- and short-term memory banks with persona modeling for sustained interaction, while RMM improves personalized retrieval through prospective and retrospective reflection over long-horizon dialogue history \cite{li2025helloagainllmpoweredpersonalized,tan2025prospectretrospectreflectivememory}.

Recent system-level research redefines memory as a standalone infrastructure layer rather than a single retrieval module. MemOS partitions memory systems into modular storage, update, retrieval and generation components, and advocates a unified OS-style abstraction to manage heterogeneous memory resources \cite{li2025memos}. Concurrent survey literature also highlights that memory systems should be evaluated based on core primitives including consolidation, update, indexing, forgetting and retrieval, rather than merely framed as simple context window extensions \cite{du2025rethinking}. Existing studies collectively frame memory design as a joint optimization problem of storage layout, record organization and retrieval strategy. Nevertheless, nearly all prior work operates under a single implicit assumption. Retrieval only needs to prioritize task-relevant memories. This assumption fails in multi-user shared agent environments, where the system must additionally validate whether candidate memories belong to the current requesting user before reasoning.

\subsection{A-Mem}
Among existing structured memory frameworks, A-Mem bears the closest connection to our bio-memory pipeline, as it organizes long-term memory into graph-connected atomic notes instead of flat unstructured dialogue archives \cite{xu2025mem}. Drawing inspiration from Zettelkasten note-taking systems, each memory unit stores complete metadata such as interaction text, timestamps, keywords, tags, descriptive context, semantic embeddings, and bidirectional links to related records. Instead of appending raw dialogue text naively when new interactions arrive, A-Mem invokes the LLM to generate structured attributes for new notes, matches them against historical records to identify semantic connections, and embeds new entries into a dynamically evolving memory graph via dedicated indexing and link construction logic. This transforms memory storage from passive log recording into an active, structured organization process.

A core highlight of A-Mem is its support for continuous memory evolution \cite{xu2025mem}. Incoming new observations trigger iterative updates to historical notes, enriching them with supplementary context, refined tag labels and strengthened cross-note links over time. Unlike prior flat memory systems that treat stored text as static unmodifiable chunks, A-Mem’s graph architecture enables lifelong refinement of memory representations as user experience accumulates. This design delivers strong performance on memory-intensive QA benchmarks. Retrieval operates over semantically enriched, interconnected note units, allowing downstream reasoning to recover not only isolated facts but complete contextual evidence chains.

A-Mem’s modular note structure makes it a natural base for our identity-aware extension. Since each note functions as an independent retrieval unit, we can attach biometric identity metadata without rewriting its core note creation, graph linking or semantic ranking modules. However, A-Mem’s retrieval logic solely relies on semantic similarity ranking, which fits single-user scenarios perfectly. The only retrieval target is matching personal history from one fixed user. In multi-user shared memory pools, semantically identical notes from different users coexist and create a critical unaddressed security risk. Semantic ranking cannot distinguish cross-user private records. To resolve this limitation, bio-memory builds upon A-Mem’s graph note architecture and inserts a biometric identity gating filter prior to semantic retrieval, rather than overhauling its fundamental memory organization design.

\subsection{Identity and Biometrics in Multi-User Agent Settings}
%讲常用的身份识别方法
As LLM agents move into shared environments, logical credentials or session identifiers no longer reliably specify the physical requester \cite{yang2026multi,lee2025spectrum}. Family assistants, public kiosks, and enterprise terminals may all serve multiple people within the same active session, so account-level identity and physical identity can diverge. Prior work has identified this ambiguity as a central challenge for multi-user agents \cite{yang2026multi} and has shown that memory systems may expose stored personal information under mismatched or crafted queries \cite{wang2025unveiling}. Existing multi-user agent research therefore highlights the need to align the active requester with the correct personalized context, but generally does not use physical identity evidence as the primary signal for constructing the retrieval candidate set.

Biometric recognition offers a direct way to ground memory access in the current requester. Face recognition methods such as ArcFace \cite{deng2019arcface} provide strong non-contact verification, while palmprint recognition captures complementary structural and textural cues and remains reliable across different sensing conditions \cite{yang2023comprehensive}. Although biometrics are commonly used for login or session authentication, prior work on continuous authentication shows that one-time identity verification is often insufficient once the active user can change during ongoing interaction \cite{li2024mbbfauth}. However, biometric evidence is still less often integrated into the retrieval pipeline itself. Bio-Memory addresses this gap by attaching biometric embeddings to memory notes and using biometric matching to determine which memories are eligible for semantic retrieval.

This distinction between \emph{session authentication} and \emph{retrieval conditioning} matters. A user may be authenticated at system entry, yet the memory layer must still decide which subset of notes should be considered for a query, especially when sessions are shared or persistent. By moving biometric evidence to the retrieval stage, Bio-Memory uses identity not only as an access credential but also as a constraint on memory search.

\section{Methodology}
\label{sec:method}

\subsection{Overview}
%先将整体方法是怎么设计的，然后我们的图说明了什么，然后将我们的2个setting：owner和non-owner
Bio-Memory is a biometric-aware memory architecture for personalized LLM agents. Built on top of A-Mem~\cite{xu2025mem}, it augments each personal memory record with a biometric embedding, so that memory retrieval depends on both semantic similarity and biometric matching. The framework preserves the original agentic memory capabilities of A-Mem while introducing identity-conditioned retrieval at query time.

\begin{figure*}[t]
\begin{center}
   \includegraphics[width=0.95\textwidth]{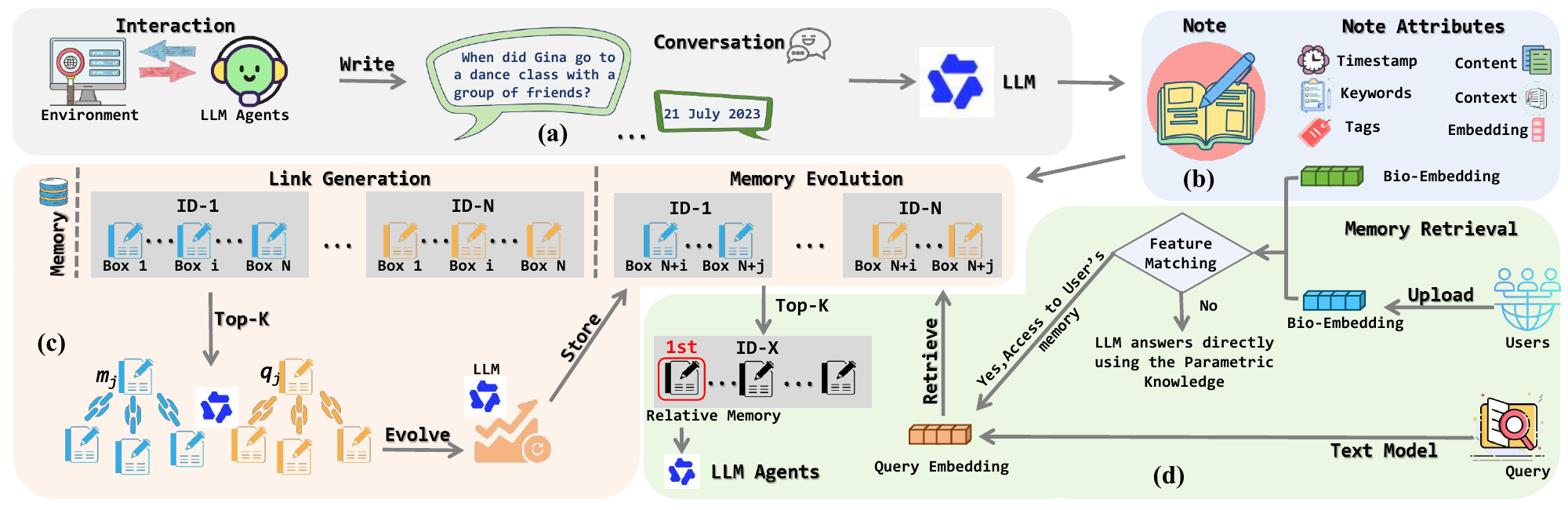}
\end{center}
\caption{Overall pipeline of Bio-Memory. (a) \textbf{Note Construction}: Raw agent-environment interactions are processed into structured memory notes with semantic attributes. (b) \textbf{Note Storage}: Each note is stored with full attributes including temporal information, textual content, contextual details, keywords, tags, semantic embeddings and biometric embeddings to support dual semantic and biometric memory management. (c) \textbf{Link Generation and Memory Evolution}: The LLM builds semantic links between memories and dynamically updates attributes of historical memories via new memory inputs, enabling continuous knowledge evolution and interconnected memory boxes. (d) \textbf{Memory Retrieval}: Biometric matching is performed prior to semantic ranking to filter valid memory candidates. Qualified memories are fed into the LLM for personalized response generation, while the model directly leverages parametric knowledge if no valid memories are available.}
\label{fig:pipeline}
\end{figure*}

Figure~\ref{fig:pipeline} illustrates the full Bio-Memory pipeline, including note construction, note storage with biometric embeddings, memory evolution, and biometric-aware retrieval.

Bio-Memory is evaluated under two query-time roles within the shared-agent protocol. In the \textbf{Owner} condition, the probe biometric embedding matches the stored biometric embedding attached to the target memory records, so those memories participate in retrieval. In the \textbf{Non-owner} condition, the probe biometric embedding does not match, so those memories are excluded from the matched candidate pool. For each question, Bio-Memory constructs a matched memory pool, sorts it chronologically, ranks it by semantic similarity, selects the top-$K$ results, and passes them to the LLM. If matching leaves no available memory, the model receives an empty memory context. Bio-Memory is therefore a lightweight retrieval-layer extension rather than a new end-to-end agent stack.

\subsection{Memory Construction}
Following the design of A-Mem, for each interaction input, Bio-Memory first constructs a structured memory note $m_i$ with rich semantic attributes. We extend this structure by incorporating a biometric embedding $b_i$ for personalization, yielding the unified representation:
\begin{equation}
m_i = \left\{ c_i, t_i, K_i, G_i, X_i, e_i, L_i, b_i \right\},
\end{equation}
where $c_i$ denotes the original interaction content, $t_i$ the timestamp, $K_i$ LLM-generated keywords, $G_i$ LLM-generated tags, $X_i$ the LLM-produced contextual description, $e_i$ the semantic embedding of the memory note, $L_i$ the set of semantically linked memories, and $b_i$ the stored biometric embedding for this memory record.

These biometric embeddings are extracted via a pre-trained biometric encoder. In our evaluation setup, all memory records from the same conversational sample are assigned to a single user $u$, sharing the same stored biometric template $b_i = b_u^{\mathrm{gal}}$. For face-based experiments, we use ArcFace; for palmprint-based experiments, we use CCNet. All biometric embeddings are L2-normalized and matched with cosine similarity.

To construct the structured attributes $K_i$, $G_i$, and $X_i$, we prompt the LLM with a dedicated template $P_{s1}$:
\begin{equation}
K_i, G_i, X_i \leftarrow \mathrm{LLM}\left(c_i \parallel t_i \parallel P_{s1}\right).
\end{equation}

The semantic embedding $e_i$ is computed by encoding the concatenation of all textual components using a text encoder $f_{\mathrm{enc}}$:
\begin{equation}
e_i = f_{\mathrm{enc}}\left(\mathrm{concat}\left(c_i, K_i, G_i, X_i\right)\right).
\end{equation}
Bio-Memory otherwise retains the original A-Mem mechanisms for link generation and memory evolution without additional modifications.

\subsection{Memory Retrieval}
Unlike conventional semantic-only retrieval, Bio-Memory conditions the retrieval candidate set on biometric matching before assessing semantic similarity. At query time step $t$, the current requester submits a query $q_t$ alongside their real-time probe biometric embedding, denoted as $b_t$.

Bio-Memory first performs online biometric matching against the memory store to determine whether the probe biometric embedding matches the stored biometric embedding of each memory record. The biometric matching output for a specific memory record $m_i$ is formalized as an indicator function:
\begin{equation}
M_{\mathrm{bio}}(b_t, b_i) = \mathbb{I}\big[\mathrm{sim}_{\mathrm{bio}}(b_t, b_i) \ge \tau_{\mathcal{D}}\big],
\end{equation}
where $\mathrm{sim}_{\mathrm{bio}}(\cdot)$ computes the biometric similarity (e.g., cosine similarity), and $\tau_{\mathcal{D}}$ is a predefined verification threshold optimized on the respective biometric benchmark $\mathcal{D}$.
In the experimental protocol, $\tau_{\mathcal{D}}$ is selected independently for each biometric benchmark based on its verification pairs. Specifically, cosine similarity scores are first computed for owner and non-owner comparisons, and $\tau_{\mathcal{D}}$ is defined as the intersection of the corresponding score distributions that lies between their principal modes. The resulting benchmark-specific threshold is then fixed for all subsequent LoCoMo Owner and Non-owner trials on that benchmark.

Memories with $M_{\mathrm{bio}} = 1$ form a query-specific matched memory pool:
\begin{equation}
\mathcal{M}^{\mathrm{match}}_t = \{m_i \in \mathcal{M} : M_{\mathrm{bio}}(b_t, b_i) = 1\}.
\end{equation}
Because this pool is constructed dynamically, personalization is determined directly by the requester's biometric embedding at query time.

This order of operations is central to the method. If semantic ranking were applied before biometric filtering, notes from different users could already compete in the same candidate set. By matching biometrics first, Bio-Memory narrows the search space before relevance ranking and makes identity consistency easier to maintain.

Subsequently, we sort the matched pool chronologically to preserve temporal coherence:
\begin{equation}
\mathcal{M}^{\mathrm{time}}_t = \operatorname{SortTime}\left(\mathcal{M}^{\mathrm{match}}_t\right).
\end{equation}

Semantic retrieval is then performed within the time-sorted pool. We encode the query $q_t$ into $e_t$ using the text encoder $f_{\mathrm{enc}}$. The semantic similarity between the query and a candidate memory is computed as cosine similarity:
\begin{equation}
\mathrm{sim}_{\mathrm{sem}}(q_t, m_i) = \frac{e_t^\top e_i}{\|e_t\|_2 \|e_i\|_2}.
\end{equation}

Bio-Memory selects the top-$K$ most relevant memories from the time-sorted pool, denoted as $\mathcal{R}_t = \operatorname{TopK}_{m_i\in\mathcal{M}^{\mathrm{time}}_t} \mathrm{sim}_{\mathrm{sem}}(q_t, m_i)$. The value of $K$ is fixed to 10 across both Owner and Non-owner conditions, ensuring that the probe biometric embedding is the only experimental difference between the two settings. The final retrieved memory set is formally defined as:
\begin{equation}
\mathcal{R}_t =
\begin{cases}
\operatorname{TopK}_{m_i\in\mathcal{M}^{\mathrm{time}}_t} \mathrm{sim}_{\mathrm{sem}}(q_t, m_i), & \mathcal{M}^{\mathrm{match}}_t \neq \emptyset,\\
\emptyset, & \mathcal{M}^{\mathrm{match}}_t = \emptyset.
\end{cases}
\end{equation}

Bio-Memory combines biometric matching, chronological ordering, and semantic ranking to determine which memories are passed to the LLM. Users asking the same question can therefore be routed to entirely different personalized memory sets.

This formulation also clarifies the role of an empty matched set. When $\mathcal{M}^{\mathrm{match}}_t = \emptyset$, the system withholds user-specific memory because the identity condition is unmet. In shared-agent settings, this is preferable to retrieving semantically plausible but identity-mismatched memories.

\subsection{Response Generation}
The retrieved memory records are serialized into a temporary context:
\begin{equation}
C_t = \mathrm{Format}(\mathcal{R}_t),
\end{equation}
and the final answer is generated by the LLM conditioned on the question and the retrieved memory context:
\begin{equation}
\hat{y}_t = \mathrm{LLM}(q_t, C_t).
\end{equation}

When $\mathcal{R}_t=\emptyset$, the formatted context is empty, so Bio-Memory falls back to answering the question without personal memory support. This behavior is particularly important in the Non-owner setting: personal memories remain entirely absent from retrieval, while the model may still generate a reasonable response based only on its parametric knowledge.

\begin{table}[t]
\centering
\small
\caption{Face recognition evaluation datasets.}
\label{tab:face-datasets}
\begin{tabular}{lcccc}
\toprule
Dataset & Protocol / Subset & Type / Group & \#Subjects & \#Images \\
\midrule
AgeDB-30~\cite{moschoglou2017agedb} & AgeDB-30 & Age & 568 & 16,488 \\
CALFW~\cite{zheng2017cross} & CALFW & Age & 5,749 & 12,174 \\
CFP-FF~\cite{sengupta2016frontal} & CFP-FF & Frontal & 500 & 7,000 \\
CFP-FP~\cite{sengupta2016frontal} & CFP-FP & Pose & 500 & 7,000 \\
CPLFW~\cite{zheng2018cross} & CPLFW & Pose & 5,749 & 11,652 \\
LFW~\cite{huang2008labeled} & LFW & Frontal & 5,749 & 13,233 \\
VGG2-FP~\cite{cao2018vggface2} & VGG2-FP & Pose & 9,131 & $\sim$3.31M \\
\bottomrule
\end{tabular}
\end{table}

\begin{table}[t]
\centering
\small
\caption{Palmprint recognition evaluation datasets.}
\label{tab:palm-datasets}
\begin{tabular}{lcccc}
\toprule
Dataset & Protocol / Subset & Type / Group & \#Subjects & \#Images \\
\midrule
CasiaM~\cite{sun2005ordinal} & \textit{CasiaM\_460} & 460nm & 200 & 1,200 \\
CasiaM~\cite{sun2005ordinal} & \textit{CasiaM\_700} & 700nm & 200 & 1,200 \\
CasiaM~\cite{sun2005ordinal} & \textit{CasiaM\_850} & 850nm & 200 & 1,200 \\
IITD~\cite{kumar2008incorporating} & IITD & Contactless & 460 & 2,300 \\
MS~\cite{zhang2009online} & \textit{MS\_Blue} & Blue & 500 & 6,000 \\
MS~\cite{zhang2009online} & \textit{MS\_Green} & Green & 500 & 6,000 \\
MS~\cite{zhang2009online} & \textit{MS\_NIR} & NIR & 500 & 6,000 \\
MS~\cite{zhang2009online} & \textit{MS\_Red} & Red & 500 & 6,000 \\
PolyU~\cite{zhang2003online} & PolyU & Contact & 378 & 7,560 \\
Tongji~\cite{zhang2017towards} & Tongji & Contactless & 600 & 12,000 \\
\bottomrule
\end{tabular}
\end{table}

\section{Experimental Setup}

\subsection{Datasets}
We evaluate Bio-Memory using LoCoMo~\cite{maharana2024evaluating} together with public biometric matching benchmarks. The face and palmprint benchmarks are summarized in Tables~\ref{tab:face-datasets} and~\ref{tab:palm-datasets}. Each LoCoMo conversational sample is paired with one stored biometric embedding, and all memory entries from that sample share the same biometric template. For each biometric benchmark, we build independent same-identity and different-identity pairings between LoCoMo samples and biometric identities.

\subsection{Evaluation Protocol}
Our goal is to test whether biometric personalization at query time can isolate personal memory retrieval in a shared-agent setting. Since LoCoMo does not include biometrics, we bind each LoCoMo conversation to one stored biometric embedding from a biometric benchmark and store all memory entries from that conversation with the same template.

To simulate multi-user usage, we construct a \textbf{10-user shared-agent} memory store by merging ten LoCoMo conversations into a single memory space containing ten biometric-linked user partitions. Each conversation is assigned one biometric identity from the corresponding benchmark, and all memories from that conversation share the same stored gallery template. For a target partition, the \textbf{Owner} query uses a probe embedding sampled from the same biometric identity as the stored gallery template, while the \textbf{Non-owner} query uses a probe embedding from a different user within the same shared setting. Biometric matching is applied before semantic retrieval, so only matched memories enter the retrieval candidate pool. If no entry satisfies biometric matching, the model answers without personal memory support.

For evaluation, we report \textbf{Multi-Hop}, \textbf{Temporal}, and \textbf{Single-Hop} questions, which directly test memory retrieval and memory-grounded reasoning. To reduce evaluation cost, we use a fixed 10\% subset of the LoCoMo test split by setting the sampling ratio to 0.1. Within each run, the same sampled question subset is reused across biometric datasets and across the \textbf{Owner} and \textbf{Non-owner} conditions, so the probe biometric embedding remains the primary experimental difference. We compare model predictions with the LoCoMo reference answers using token-level F1 and BLEU-1. All scores are reported as percentages, the reported Average is the macro-average over the three question types, and Avg. Gap is computed as Owner Average minus Non-owner Average.

We emphasize these three question types because they most directly reflect the effect of retrieval quality on downstream answering. Single-Hop questions test whether one key fact can be recovered from the correct personal memory partition. Multi-Hop questions require the model to combine multiple pieces of stored evidence, making them more sensitive to noisy or identity-misaligned retrieval. Temporal questions are especially informative because they depend on the ordering of personal events and are difficult to answer from generic parametric knowledge alone.

\subsection{Implementation Details}
Our implementation builds upon the A-Mem memory layer and adds biometric personalization at retrieval time. Semantic indexing uses a sentence-transformer retriever based on \texttt{all-MiniLM-L6-v2}, while the answer generation model and inference backend follow the evaluation script. To ensure fair comparison, owner and non-owner trials share the same memory store, retrieval configuration, and evaluation subset; the only inference difference is the query probe biometric embedding.

\begin{table*}[t]
\centering
\small
\setlength{\tabcolsep}{4.2pt}
\caption{Downstream LoCoMo QA results under palmprint-based retrieval personalization in the 10-user shared-agent setting. Each entry is reported as F1 / BLEU-1 in percentages (\%). Average is the macro-average over Multi-Hop, Temporal, and Single-Hop questions, and Avg. Gap is Owner Average minus Non-owner Average.}
\label{tab:palm-main-results}
\resizebox{\textwidth}{!}{%
\begin{tabular}{llccccc}
\toprule
\textbf{Dataset} & \textbf{Condition} & \textbf{Multi-Hop} & \textbf{Temporal} & \textbf{Single-Hop} & \textbf{Average} & \textbf{Avg. Gap} \\
\midrule
\multirow{2}{*}{\textbf{PolyU}}
& \textbf{Owner}    & 23.70 / 16.32 & 31.16 / 21.19 & 27.08 / 22.67 & 27.31 / 20.06 & \multirow{2}{*}{22.82 / 16.18} \\
& \textbf{Non-owner} & 5.26 / 4.67 & 0.36 / 0.27 & 7.84 / 6.71 & 4.49 / 3.88 & \\
\midrule
\multirow{2}{*}{\textbf{IITD}}
& \textbf{Owner}    & 20.64 / 17.79 & 23.73 / 16.10 & 34.43 / 29.85 & 26.27 / 21.25 & \multirow{2}{*}{20.75 / 16.75} \\
& \textbf{Non-owner} & 5.27 / 4.84 & 3.09 / 1.66 & 8.21 / 7.00 & 5.52 / 4.50 & \\
\midrule
\multirow{2}{*}{\textbf{CasiaM\_850}}
& \textbf{Owner}    & 20.41 / 16.55 & 28.14 / 18.77 & 31.65 / 25.41 & 26.73 / 20.24 & \multirow{2}{*}{20.13 / 14.97} \\
& \textbf{Non-owner} & 7.51 / 6.07 & 3.45 / 1.93 & 8.85 / 7.80 & 6.60 / 5.27 & \\
\midrule
\multirow{2}{*}{\textbf{MS\_Green}}
& \textbf{Owner}    & 26.46 / 21.09 & 29.58 / 21.24 & 34.49 / 29.42 & 30.18 / 23.92 & \multirow{2}{*}{24.47 / 19.11} \\
& \textbf{Non-owner} & 5.62 / 5.16 & 2.82 / 1.49 & 8.70 / 7.79 & 5.71 / 4.81 & \\
\midrule
\multirow{2}{*}{\textbf{CasiaM\_460}}
& \textbf{Owner}    & 25.71 / 19.86 & 31.71 / 21.86 & 36.53 / 30.28 & 31.32 / 24.00 & \multirow{2}{*}{25.73 / 19.13} \\
& \textbf{Non-owner} & 5.72 / 6.35 & 3.56 / 2.06 & 7.48 / 6.19 & 5.59 / 4.87 & \\
\midrule
\multirow{2}{*}{\textbf{MS\_Red}}
& \textbf{Owner}    & 19.62 / 17.50 & 30.70 / 19.89 & 33.45 / 28.11 & 27.92 / 21.83 & \multirow{2}{*}{21.58 / 16.82} \\
& \textbf{Non-owner} & 5.83 / 4.88 & 3.20 / 1.79 & 9.99 / 8.36 & 6.34 / 5.01 & \\
\midrule
\multirow{2}{*}{\textbf{CasiaM\_700}}
& \textbf{Owner}    & 25.72 / 20.32 & 27.48 / 18.90 & 34.38 / 29.69 & 29.19 / 22.97 & \multirow{2}{*}{24.47 / 18.59} \\
& \textbf{Non-owner} & 5.39 / 5.39 & 1.83 / 1.39 & 6.95 / 6.37 & 4.72 / 4.38 & \\
\midrule
\multirow{2}{*}{\textbf{MS\_NIR}}
& \textbf{Owner}    & 27.50 / 22.80 & 30.35 / 20.61 & 31.53 / 27.96 & 29.79 / 23.79 & \multirow{2}{*}{24.52 / 19.53} \\
& \textbf{Non-owner} & 2.88 / 3.07 & 3.48 / 1.96 & 9.46 / 7.75 & 5.27 / 4.26 & \\
\midrule
\multirow{2}{*}{\textbf{MS\_Blue}}
& \textbf{Owner}    & 22.26 / 16.97 & 35.89 / 24.28 & 33.90 / 29.43 & 30.68 / 23.56 & \multirow{2}{*}{25.75 / 19.22} \\
& \textbf{Non-owner} & 5.61 / 5.48 & 2.43 / 1.19 & 6.76 / 6.35 & 4.93 / 4.34 & \\
\midrule
\multirow{2}{*}{\textbf{Tongji}}
& \textbf{Owner}    & 23.21 / 18.20 & 29.10 / 20.12 & 29.89 / 24.63 & 27.40 / 20.98 & \multirow{2}{*}{22.19 / 16.68} \\
& \textbf{Non-owner} & 4.31 / 4.70 & 4.23 / 1.95 & 7.08 / 6.25 & 5.21 / 4.30 & \\
\bottomrule
\end{tabular}}
\end{table*}

\begin{table*}[t]
\centering
\small
\setlength{\tabcolsep}{4.2pt}
\caption{Downstream LoCoMo QA results under face-based retrieval personalization in the 10-user shared-agent setting. Each entry is reported as F1 / BLEU-1 in percentages (\%). Average is the macro-average over Multi-Hop, Temporal, and Single-Hop questions, and Avg. Gap is Owner Average minus Non-owner Average.}
\label{tab:face-main-results}
\resizebox{\textwidth}{!}{%
\begin{tabular}{llccccc}
\toprule
\textbf{Dataset} & \textbf{Condition} & \textbf{Multi-Hop} & \textbf{Temporal} & \textbf{Single-Hop} & \textbf{Average} & \textbf{Avg. Gap} \\
\midrule
\multirow{2}{*}{\textbf{AgeDB-30}}
& \textbf{Owner}    & 22.99 / 15.60 & 36.46 / 25.70 & 34.77 / 27.90 & 31.41 / 23.07 & \multirow{2}{*}{26.85 / 19.58} \\
& \textbf{Non-owner} & 3.22 / 1.88 & 1.38 / 1.03 & 9.07 / 7.55 & 4.56 / 3.49 & \\
\midrule
\multirow{2}{*}{\textbf{CALFW}}
& \textbf{Owner}    & 23.16 / 20.15 & 33.92 / 23.02 & 35.58 / 29.77 & 30.89 / 24.31 & \multirow{2}{*}{27.29 / 21.15} \\
& \textbf{Non-owner} & 2.04 / 2.19 & 2.21 / 1.11 & 6.56 / 6.17 & 3.60 / 3.16 & \\
\midrule
\multirow{2}{*}{\textbf{CFP-FF}}
& \textbf{Owner}    & 21.17 / 18.85 & 31.81 / 22.27 & 37.69 / 31.11 & 30.22 / 24.08 & \multirow{2}{*}{26.34 / 20.72} \\
& \textbf{Non-owner} & 1.92 / 1.49 & 1.47 / 1.12 & 8.26 / 7.48 & 3.88 / 3.36 & \\
\midrule
\multirow{2}{*}{\textbf{CFP-FP}}
& \textbf{Owner}    & 24.25 / 19.85 & 32.42 / 21.30 & 37.16 / 31.12 & 31.28 / 24.09 & \multirow{2}{*}{26.18 / 19.32} \\
& \textbf{Non-owner} & 5.40 / 5.45 & 1.08 / 0.82 & 8.81 / 8.03 & 5.10 / 4.77 & \\
\midrule
\multirow{2}{*}{\textbf{CPLFW}}
& \textbf{Owner}    & 20.54 / 17.61 & 31.27 / 21.58 & 33.96 / 28.52 & 28.59 / 22.57 & \multirow{2}{*}{23.58 / 18.34} \\
& \textbf{Non-owner} & 4.47 / 3.99 & 2.36 / 1.42 & 8.19 / 7.28 & 5.01 / 4.23 & \\
\midrule
\multirow{2}{*}{\textbf{LFW}}
& \textbf{Owner}    & 21.77 / 18.23 & 36.53 / 27.31 & 32.96 / 27.65 & 30.42 / 24.40 & \multirow{2}{*}{25.81 / 20.60} \\
& \textbf{Non-owner} & 2.19 / 2.21 & 2.73 / 1.39 & 8.91 / 7.80 & 4.61 / 3.80 & \\
\midrule
\multirow{2}{*}{\textbf{VGG2-FP}}
& \textbf{Owner}    & 22.28 / 18.58 & 31.88 / 21.57 & 38.49 / 33.10 & 30.88 / 24.42 & \multirow{2}{*}{25.64 / 19.98} \\
& \textbf{Non-owner} & 5.85 / 5.06 & 3.12 / 1.69 & 6.75 / 6.57 & 5.24 / 4.44 & \\
\bottomrule
\end{tabular}}
\end{table*}

\section{Results}
\subsection{Quantitative Results}
Tables~\ref{tab:palm-main-results} and~\ref{tab:face-main-results} summarize the quantitative results on palmprint and face benchmarks in the 10-user shared-agent setting. Across both tables, the same pattern appears repeatedly: once retrieval is conditioned on biometric identity, owner queries preserve access to the correct personal evidence, whereas non-owner queries lose that evidence before semantic ranking begins.

\noindent\textbf{Overall trend.} The owner advantage is uniform across all datasets, all question types, and both evaluation metrics. This gap is not small. In the palmprint results, the average owner--non-owner gap ranges from 20.13 to 25.75 in F1 and from 14.97 to 19.53 in BLEU-1; the strongest separation appears on MS\_Blue, where the owner average reaches 30.68 / 23.56 while the non-owner average drops to 4.93 / 4.34. The face benchmarks show the same behavior, with average gaps from 23.58 to 27.29 in F1 and from 18.34 to 21.15 in BLEU-1. CALFW is the clearest example: the owner average is 30.89 / 24.31, compared with only 3.60 / 3.16 for the non-owner. Because the owner and non-owner settings use the same memory store, the same question subset, and the same downstream generator, these consistent margins indicate that the decisive change happens at the biometric filtering stage.

\noindent\textbf{Authorized utility.} The owner and non-owner rows also make clear that the biometric filtering step is not achieved by sacrificing useful memory for the legitimate user. In the palmprint table, owner averages remain stable between 26.27 / 21.25 and 31.32 / 24.00 across all ten protocols, whereas non-owner averages stay much lower, mostly between 4.49 / 3.88 and 6.60 / 5.27. The face table exhibits the same structure: owner averages stay in a narrow and healthy band from 28.59 / 22.57 to 31.41 / 23.07, while non-owner averages remain near the floor, between 3.60 / 3.16 and 5.24 / 4.44. This contrast matters for interpretation. If biometric conditioning simply removed too much memory from everyone, both rows in each dataset would collapse together. Instead, the owner rows remain strong, which means the filtering step is selective: it suppresses identity-mismatched evidence while preserving enough user-aligned notes for effective memory-grounded answering.

\noindent\textbf{Question-type behavior.} The per-task columns further explain where the separation comes from. Temporal questions are the most sensitive. In the palmprint results, owner Temporal F1 scores range from 23.73 to 35.89, but the corresponding non-owner scores are only 0.36 to 4.23; in the face results, owner Temporal F1 remains between 31.81 and 36.53, whereas non-owner Temporal F1 stays between 1.08 and 3.12. This is consistent with the nature of Temporal QA: once the correct user's event chain is removed, the model has little basis for reconstructing the timeline. The Single-Hop and Multi-Hop columns show a related but slightly different pattern. Single-Hop owner scores stay strong in both tables because one key personal fact is often recoverable when the correct partition is retained, while Multi-Hop owner scores remain competitive because several related notes can still be retrieved together. The non-owner rows, by contrast, stay low across both columns, indicating that biometric filtering breaks not only isolated fact access but also evidence composition.

\noindent\textbf{Cross-modal consistency.} This effect is not tied to one biometric modality. Palmprint protocols and face benchmarks differ in sensing conditions, feature variation, and verification difficulty, yet both result tables preserve the same directional pattern: owner scores remain high and non-owner scores remain suppressed. The main difference lies in the sharpness of the margin. For example, the face benchmarks reach a 27.29 / 21.15 average gap on CALFW, while the palmprint protocols reach 25.75 / 19.22 on MS\_Blue. Even the weaker cases still maintain large separations, such as 20.13 / 14.97 on PolyU and 23.58 / 18.34 on CPLFW. This consistency across modalities supports the same interpretation: better verification produces a sharper memory boundary, but even when verification is less favorable, the identity-aware retrieval mechanism still preserves a clear distinction between matched and mismatched users.

\noindent\textbf{Main takeaway.} Taken together, these results show that Bio-Memory is not merely improving generic semantic retrieval. The crucial intervention happens earlier, at the stage where the candidate pool is constructed. Across all rows, owner queries retain access to the personalized notes needed for high-quality answers, while non-owner queries are consistently restricted to incomplete or generic context. The fact that this pattern holds simultaneously across palmprint protocols, face benchmarks, and all three question types indicates that the method acts as retrieval-time access control over memory evidence rather than as a post hoc adjustment of generated responses.
\begin{figure*}[t]
\centering
\includegraphics[width=0.95\textwidth]{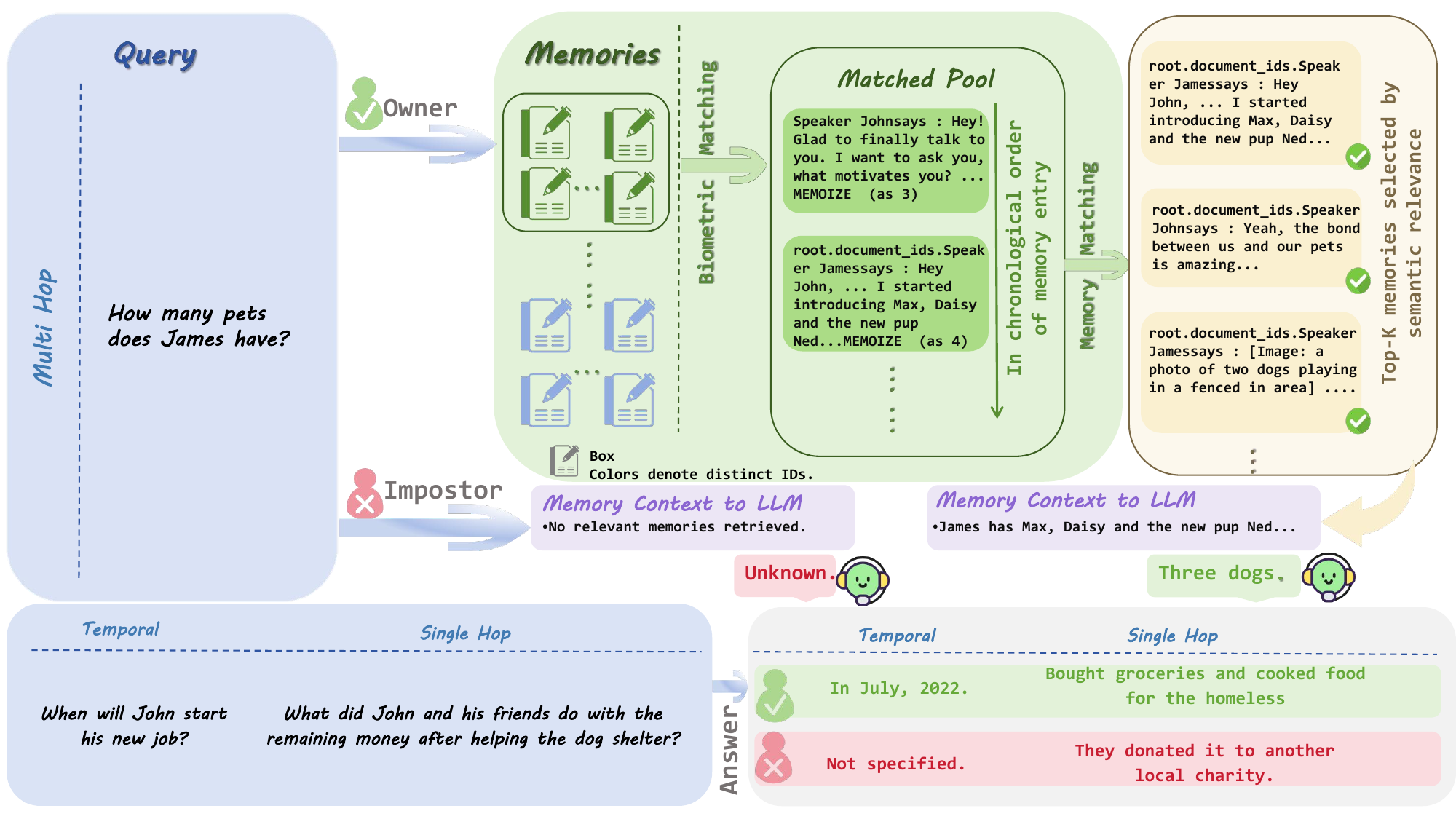}
\caption{Retrieval traces for owner-versus-non-owner queries in Bio-Memory. For the same question, owner and non-owner queries follow different retrieval paths because their probe biometric embeddings align differently with the personal memory store.}
\label{fig:qa}
\end{figure*}

\subsection{Qualitative Retrieval Analysis}

\noindent\textbf{Qualitative examples.} Figure~\ref{fig:qa} gives concrete examples of the same effect. In the Multi-Hop case, the owner retrieves the right notes and answers \emph{``Three dogs,''} while the non-owner falls back to \emph{``Unknown.''} In the Temporal case, the owner answers \emph{``In July, 2022,''} while the non-owner responds \emph{``Not specified.''} In the Single-Hop case, the owner retrieves the relevant personal fact, whereas the non-owner produces an unsupported alternative. These examples are easy to read because they show the same question under the same system with only one change: the identity signal used at retrieval time.

\noindent\textbf{Interpretation.} The qualitative traces confirm that the main separation happens before answer generation. Owner and non-owner queries diverge as soon as candidate memories are selected. Once the correct notes are filtered out, Multi-Hop reasoning loses its evidence chain, Temporal questions lose the user-specific timeline, and even Single-Hop questions lose the one fact they need. The figure therefore supports the same conclusion as the tables: Bio-Memory works by controlling access to user-specific memory evidence at retrieval time.

\section{Conclusion}
This paper presented Bio-Memory, a biometric-aware memory architecture for personalized LLM agents that conditions personal memory retrieval on both semantic relevance and biometric matching at query time. By augmenting each memory note with a stored biometric embedding, Bio-Memory enables a shared agent to construct user-aligned retrieval contexts without changing the underlying LLM. Across seven face benchmarks, ten palmprint protocols, and a 10-user shared-agent evaluation, owners retain strong memory-grounded question answering performance, whereas non-owners lose access to the user-specific evidence needed for retrieval-based answering. These results support the feasibility and practical value of biometrics as a control signal for personalized memory retrieval in shared-agent settings.

More broadly, the study suggests that personalized memory for LLM agents is not only a relevance problem but also a user-alignment problem: the system must determine not only \emph{what} memory is relevant, but also \emph{whose} memory should be retrieved. Bio-Memory offers a practical way to introduce this distinction while preserving the strengths of structured memory systems such as A-Mem.

\bibliographystyle{unsrt}
\bibliography{main}
\end{document}